\documentclass[runningheads]{llncs}
\usepackage[T1]{fontenc}
\usepackage{graphicx}
\usepackage{booktabs}
\usepackage{hyperref}
\usepackage{seqsplit}
\usepackage{url}
\begin{document}
\title{moBERTo: A Modern Encoder for Portuguese via Continued Pretraining of ModernBERT}
%
%
\author{Thiago Laitz\inst{1,2,3} \and
Thales Sales Almeida\inst{1,2,3} \and \\
João Guilherme Alves Santos\inst{1,2} \and
Giovana Kerche Bonás\inst{1,2,3}}
\authorrunning{T. Laitz et al.}
\institute{UNICAMP, Campinas, Brazil \and
Tropic AI \and
Maritaca AI\\
\email{thiagolaitz@gmail.com}}

%
\maketitle
\begin{abstract}
Encoder-only transformer models remain essential for production NLP pipelines. We introduce moBERTo, a Portuguese adaptation of ModernBERT obtained through continued pretraining of the ModernBERT-base checkpoint on 60 billion tokens (5 epochs over a 12-billion-token corpus curated from FineWeb2 and filtered with educational and STEM classifiers). We preserve the original architecture, including rotary positional embeddings, alternating local–global attention, flash attention, and unpadding. We evaluate moBERTo across information retrieval (including long-context retrieval at up to 8{,}192 tokens), document classification, named entity recognition, and natural language understanding. Our best variant, which combines a Portuguese tokenizer with subword-matching embedding transfer and long-context post-training, achieves the highest average reranking nDCG@10 across three Portuguese retrieval benchmarks and the best results on PLUE-PT. Through ablation studies, we show that (i) continued pretraining is strongly preferable to training from scratch, particularly for preserving long-context capabilities; (ii) tokenizer adaptation improves token-level tasks but degrades long-context retrieval; (iii) a dedicated long-context post-training phase at 8{,}192 tokens further improves reranking and NER; and (iv) encoder-only architectures remain competitive with larger decoder-only alternatives for discriminative tasks. We publicly release the model weights (\href{https://huggingface.co/Tropic-AI/moBERTo}{\texttt{Tropic-AI/moBERTo}}) and training data (\href{https://huggingface.co/datasets/Tropic-AI/moberto-pretraining-dataset-c4-compatible}{\texttt{\seqsplit{Tropic-AI/moberto-pretraining-dataset-c4-compatible}}}) on Hugging Face.

\end{abstract}
\section{Introduction}

Encoder-only transformer models, such as BERT~\cite{devlin2019bert} and its successors, remain the backbone of numerous natural language processing (NLP) pipelines. Despite the growing popularity of large decoder-only language models, encoders continue to be used on production deployments for tasks such as information retrieval, document classification, named entity recognition, and semantic search, owing to their favorable trade-off between performance, latency, and computational cost~\cite{modernbert}. For the English language, significant progress has been made in modernizing the encoder paradigm. ModernBERT~\cite{modernbert} introduced a series of architectural and training improvements including rotary positional embeddings (RoPE), alternating local-global attention, flash attention, and unpadding being more efficient than previous encoders.
Trained on 2 trillion tokens with a native context length of $8192$ tokens, ModernBERT represents a major improvement over older models such as RoBERTa~\cite{liu2019roberta} and DeBERTaV3~\cite{he2021debertav3}.

However, the landscape for Portuguese remains considerably less developed.
The most widely adopted monolingual encoder for Portuguese is BERTimbau~\cite{souza2020bertimbau}, a BERT model trained on the BrWaC corpus~\cite{wagner2018brwac}.
While BERTimbau demonstrated clear advantages over multilingual BERT at the time of its release, it inherits the architectural limitations of the original BERT such as a 512-token context window and absolute positional embeddings. Other efforts, such as multilingual models (mBERT, XLM-R~\cite{conneau2020unsupervised}), provide Portuguese coverage but typically underperform dedicated monolingual models on language-specific benchmarks.
As a result, the Portuguese NLP community lacks access to a more modern, efficient encoder that incorporates the advances of the past several years.

In this work, we address this gap by training \textbf{moBERTo}, a Portuguese adaptation of ModernBERT. Starting from the original ModernBERT-base checkpoint, we perform continued pretraining on a curated corpus extracted from FineWeb2~\cite{penedo2025fineweb2}, a large-scale, web dataset further filtered using the educational and STEM classifiers from ClassiCC-PT~\cite{almeida2025building}, and trained the model for around 60 billion tokens (approximately 5 epochs on the dataset). We also evaluate several strategies for adapting the tokenizer and embedding layer to better represent Portuguese, as well as a dedicated long-context post-training phase.

We conduct an evaluation of moBERTo across five families of 
Portuguese downstream tasks: information retrieval, long-context 
retrieval, document classification, named entity recognition, and 
natural language understanding. We additionally evaluate on English 
GLUE to measure language retention. Our results show that moBERTo outperforms existing Portuguese encoders, including BERTimbau, across the majority of evaluated tasks, while retaining the efficiency advantages of the ModernBERT architecture.

\noindent In summary, our contributions are as follows:
\begin{enumerate}
    \item We release \textbf{moBERTo}, a Portuguese adaptation of the ModernBERT architecture, bringing modern encoder design to Portuguese NLP.
    \item We provide a \textbf{comprehensive evaluation} across multiple Portuguese benchmarks covering retrieval, classification, NER, and similarity tasks.
    \item We conduct \textbf{ablation studies} on tokenizer adaptation, subword-matching embedding transfer, long-context post-training, and base architecture choice.
    \item We publicly release the model weights\footnote{\href{https://huggingface.co/Tropic-AI/moBERTo}{huggingface.co/Tropic-AI/moBERTo}} and training dataset\footnote{\href{https://huggingface.co/datasets/Tropic-AI/moberto-pretraining-dataset-c4-compatible}{huggingface.co/datasets/Tropic-AI/moberto-pretraining-dataset}}.
\end{enumerate}


\section{Related Work}

While multilingual models such as mBERT and XLM-R~\cite{conneau2020unsupervised} offer broad language coverage, dedicated monolingual encoders consistently outperform them on language-specific benchmarks, as demonstrated by BERTimbau \cite{souza2020bertimbau} for Portuguese, CamemBERT~\cite{martin2020camembert} for French, and PhoBERT \cite{nguyen2020phobert} for Vietnamese, a\-mong others. These models generally follow the BERT~\cite{devlin2019bert} or RoBERTa~\cite{liu2019roberta} pretraining recipes, and more recent efforts such as CamemBERT 2.0~\cite{antoun2024camembert} and PortBERT~\cite{scheible2025portbert} confirm that monolingual encoder development remains an active line of work.

In parallel, recent studies have revisited the encoder-only paradigm with targeted architectural improvements~\cite{breton2025neobert,eurobert2025,ettin2025}. ModernBERT~\cite{modernbert} introduced alternating local-global attention, sequence unpadding, and scaled pretraining to 2 trillion tokens with 8{,}192-token context, achieving state-of-the-art results on classification and retrieval benchmarks. NeoBERT~\cite{neobert2025} offers a complementary design with an optimized depth-to-width ratio and Pre-RMSNorm, reporting strong results on MTEB~\cite{muennighoff2022mteb} at 250M parameters. These modernized recipes have since been adapted to other languages, including German~\cite{moderngbert}, Japanese~\cite{sugiura2025llmjpmodernbert}, and a massively multilingual setting covering over 1{,}800 languages~\cite{mmbert2025}.

In the Portuguese NLP ecosystem, BERTimbau~\cite{souza2020bertimbau} 
remains the most widely used encoder, while Albertina PT~\cite{rodrigues2023advancing,santos-etal-2024-fostering} 
extended coverage with DeBERTa-based models at multiple scales. 
BERTugues~\cite{mazzazago2024bertugues} followed the BERTimbau recipe 
but improved the tokenizer by removing rarely used characters and adding 
over 7{,}000 Portuguese-specific tokens, reporting gains over BERTimbau 
across several downstream tasks. Domain-specific encoders have also 
appeared for legal~\cite{garcia2024robertalexpt}, biomedical~\cite{schneider2020biobertpt}, 
and governmental text~\cite{silva2024govbert}. More recent efforts have begun to bring modern encoder architectures to Portuguese. ModBERTBr~\cite{eniac} introduced a ModernBERT-inspired model trained from scratch on BrWAC and Wikipedia, while NeoBERTugues~\cite{cesconetto2026neobertugues} adapted the Modernbert architecture for Portuguese. moBERTo complements these efforts by adapting ModernBERT via continued pretraining from the original 
checkpoint on a curated FineWeb2 corpus.

\section{Methodology}
\label{sec:methodology}

\begin{figure}[t]
    \centering
    \includegraphics[width=\linewidth]{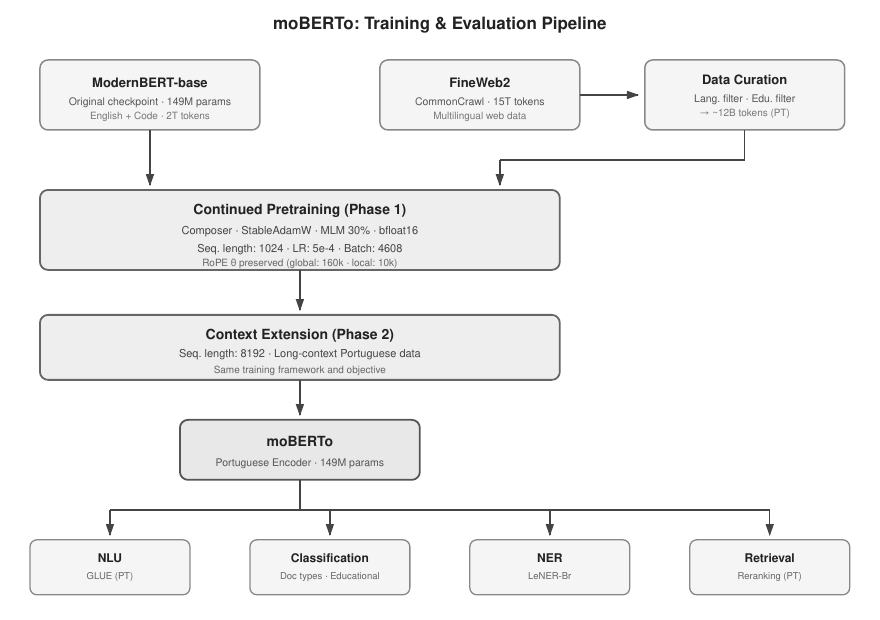}
    \caption{Overview of the moBERTo training and evaluation pipeline. We start from the original ModernBERT-base checkpoint and perform continued pretraining on 60B tokens of curated Portuguese data, followed by evaluation across multiple downstream tasks.}
    \label{fig:pipeline}
\end{figure}

Our approach consists of adapting the original ModernBERT-base 
checkpoint to Portuguese through continued pretraining on a curated 
corpus of approximately 12 billion tokens for five epochs at a 
sequence length of 1{,}024, optionally followed by a long-context 
post-training phase at 8{,}192 tokens and combined with tokenizer 
adaptation. We intentionally preserve the original training 
configuration as closely as possible, modifying only what is 
strictly necessary for language adaptation. An overview of our 
pipeline is illustrated in Fig.~\ref{fig:pipeline}.

\subsection{Data}
\label{sec:data}

We construct our pretraining corpus by curating the Portuguese subset of FineWeb2 \cite{penedo2025fineweb2}, a large-scale web dataset derived from CommonCrawl. We further filter the data using the educational and STEM classifiers from ClassiCC-PT, which have been shown to improve continued pretraining of LLMs~\cite{almeida2025building,almeida2025curi}. The resulting corpus comprises approximately 12 billion tokens, covering a broad range of domains and topics in Portuguese, and is roughly six times larger than BrWaC~\cite{wagner2018brwac}.

\subsection{Continued Pretraining}
\label{sec:pretraining}

We initialize our model from the publicly available ModernBERT-base checkpoint\footnote{\url{https://huggingface.co/answerdotai/ModernBERT-base}} and perform continued pretraining using the Composer framework.

\paragraph{Training objective.} We use the standard Masked Language Modeling (MLM) objective with a masking rate of 30\%, consistent with the original ModernBERT training recipe~\cite{modernbert}.

\paragraph{Hyperparameters.} We preserve the final training configuration reported by ModernBERT~\cite{modernbert} for the base model. Specifically, we maintain the same learning rate, batch size, and optimizer settings (StableAdamW) used in the original pretraining. We also apply a warmup phase, as we resume from a fully converged checkpoint rather than training from scratch. The RoPE frequency parameters are kept at their original values (160{,}000 for global attention layers; 10{,}000 for local attention layers). A summary of our training hyperparameters is provided in Table~\ref{tab:training_config}.

\begin{table}[t]
    \centering
    \caption{Training configuration for moBERTo continued pretraining 
    and long-context post-training. Both phases share the same 
    hyperparameters; the long-context phase differs only in sequence 
    length, batch size, and total training tokens.}
    \label{tab:training_config}
    \setlength{\tabcolsep}{12pt}
    \begin{tabular}{lcc}
        \toprule
        \textbf{Parameter} & \textbf{Continued} & \textbf{Long-context} \\
                           & \textbf{pretraining} & \textbf{post-training} \\
        \midrule
        Training tokens & 60B & 10B \\
        Max sequence length & 1024 & 8192 \\
        Batch size & 4608 & 576 \\
        Masking rate & \multicolumn{2}{c}{30\%} \\
        Optimizer & \multicolumn{2}{c}{StableAdamW} \\
        Learning rate & \multicolumn{2}{c}{5e-4} \\
        Weight decay & \multicolumn{2}{c}{1e-5} \\
        Dropout (attn output) & \multicolumn{2}{c}{0.1} \\
        Dropout (other) & \multicolumn{2}{c}{0.0} \\
        Precision & \multicolumn{2}{c}{bfloat16} \\
        Hardware & \multicolumn{2}{c}{4 $\times$ B200} \\
        Training framework & \multicolumn{2}{c}{Composer} \\
        \bottomrule
    \end{tabular}
\end{table}

\paragraph{Sequence length.} We perform the main continued pretraining 
phase exclusively at a maximum sequence length of 1{,}024 tokens. 
Although the original ModernBERT supports sequences of up to 8{,}192 
tokens through its context extension phase, we hypothesize that the 
long-context capabilities acquired during the original English 
pretraining are largely preserved through continued pretraining at 
shorter sequence lengths. We additionally explore an optional 
long-context post-training phase of 10B tokens at 8{,}192-token 
context, applied after the main pretraining run, to assess whether 
explicit long-context training in Portuguese yields further gains. 
Both hypotheses are evaluated in Sect.~\ref{sec:results}.

\subsection{Model Variants}
\label{sec:variants}

To isolate the effect of different adaptation strategies, we train several variants of moBERTo. In the main evaluation (Sect.~\ref{sec:results}), we report results for the four primary variants; additional variants used exclusively in ablation studies (Sect.~\ref{sec:ablations}) are marked below.

\begin{itemize}
    \item \textbf{moBERTo (orig.\ tok.)}: The base model using the original ModernBERT tokenizer, trained for 60B tokens at 1{,}024-token context.
    \item \textbf{moBERTo-8k (orig.\ tok.)}: Starting from the moBERTo (orig.\ tok.) checkpoint, we perform an additional post-training phase of 10B tokens at a maximum sequence length of 8{,}192 tokens, aiming to improve long-context capabilities in Portuguese.
    \item \textbf{moBERTo-SWM (PT tok.)}: Uses a Portuguese tokenizer whose vocabulary is constructed via a subword matching procedure that decomposes each new token into subwords of the original ModernBERT tokenizer. The embedding layer is reconstructed by combining the original embeddings of the matched subwords, following a subword matching (SWM) transfer approach that preserves partial alignment with the original embedding space. Trained for 60B tokens at 1{,}024-token context.
    \item \textbf{moBERTo-SWM-8k (PT tok.)}: Starting from the moBERTo-SWM checkpoint, we perform the same long-context post-training phase (10B tokens at 8{,}192 tokens).
    \item \textbf{moBERTo-Tok (PT tok., orig.\ emb.)} \textit{[ablation only]}: Uses the same Portuguese tokenizer as moBERTo-SWM (i.e., text is tokenized with the Portuguese vocabulary), but retains the original ModernBERT embedding layer without reconstruction. This means the new Portuguese token IDs index into embeddings that were learned for different (English) tokens, creating a misalignment that the model must resolve during continued pretraining.
    \item \textbf{moBERTo (scratch)} \textit{[ablation only]}: Uses the same architecture, data, and hyperparameters as moBERTo (orig.\ tok.) but is initialized from random weights rather than the pretrained checkpoint.
\end{itemize}

\subsection{Evaluation Benchmarks}
\label{sec:eval_protocol}

We evaluate moBERTo across five families of downstream tasks. For 
each task, we fine-tune the pretrained model and report standard 
metrics, comparing against monolingual and multilingual baselines.

\begin{enumerate}
    \item \textbf{Information Retrieval (IR):} Cross-encoder reranking 
    on QUATI~\cite{quati}, m\-MARCO-PT~\cite{mmarco}, and Robust04-PT~\cite{mrobust}, with nDCG@10. Rerankers are fine-tuned on mMARCO-PT triples.
    
    \item \textbf{Long-Context Retrieval:} Reranking on 
    MLDR~\cite{mldr}, a multilingual long-document retrieval 
    benchmark, evaluated at maximum sequence lengths of 512, 2{,}048, 
    4{,}096, and 8{,}192 tokens. The models evaluated were trained on mMARCO-PT triples.

    \item \textbf{Document Classification:} Two Portuguese 
    classification tasks: (i) detecting whether a document contains 
    educational content, and (ii) identifying the content type of a 
    given text (e.g., news, legal, academic). We report F1 as the default metric.

    \item \textbf{Named Entity Recognition (NER):} Token-level 
    sequence labeling on LeNER-Br~\cite{lenerbr}, reporting F1.

    \item \textbf{Natural Language Understanding (NLU):} 
    PLUE-PT~\cite{pluept}, the Portuguese translation of GLUE, 
    covering tasks such as semantic textual similarity and natural 
    language inference. We additionally report results on the 
    original English GLUE~\cite{glue} to assess the trade-off 
    between Portuguese adaptation and English retention.
\end{enumerate}

For each task family, we compare moBERTo against equivalent models.

\section{Results}
\label{sec:results}

We present our main results across all evaluation benchmarks in Tables~\ref{tab:ir_results}--\ref{tab:nlu_ner_results}. We report the four primary moBERTo variants alongside baselines and adapted models from other architectures; additional variants (moBERTo-Tok and moBERTo-scratch) are analyzed in the ablation studies (Sect.~\ref{sec:ablations}). For information retrieval benchmarks (QUATI, mMARCO-PT, Robust04-PT), all models were fine-tuned as rerankers training on mMARCO-PT triples.

\begin{table}[t]
    \centering
    \caption{Information retrieval (reranking) results, reported as nDCG@10. Best results in \textbf{bold}, second best \underline{underlined}.}
    \label{tab:ir_results}
    \begin{tabular}{lcccc}
        \toprule
        \textbf{Model} & \textbf{QUATI} & \textbf{mMARCO} & \textbf{Rob04-pt} & \textbf{Avg.} \\
        \midrule
        \multicolumn{5}{l}{\textit{Baselines}} \\
        BERT-base         & 0.2846 & 0.4050 & 0.2389 & 0.3095 \\
        BERTimbau-base    & 0.4870 & 0.5005 & 0.4138 & 0.4671 \\
        ModernBERT-base   & 0.3779 & 0.4799 & 0.2988 & 0.3855 \\
        NeoBERT-base      & 0.4000 & 0.4698 & 0.3117 & 0.3938 \\
        Qwen-base         & 0.4248 & 0.5065 & 0.2994 & 0.4102 \\
        \midrule
        \multicolumn{5}{l}{\textit{Our models}} \\
        moBERTo (orig.\ tok.)      & 0.5383 & 0.5109 & 0.4510 & 0.5001 \\
        moBERTo-8k (orig.\ tok.)   & 0.5231 & 0.5089 & 0.4516 & 0.4945 \\
        moBERTo-SWM (PT tok.)      & \underline{0.5410} & \textbf{0.5169} & \underline{0.4782} & \underline{0.5120} \\
        moBERTo-SWM-8k (PT tok.)   & \textbf{0.5609} & \underline{0.5147} & \textbf{0.5010} & \textbf{0.5255} \\
        \midrule
        \multicolumn{5}{l}{\textit{Other architectures}} \\
        NeoBERT-PT           & 0.4100 & 0.4654 & 0.3139 & 0.3964 \\
        Qwen3-0.6B-PT              & 0.4547 & 0.5144 & 0.2895 & 0.4195 \\
        \bottomrule
    \end{tabular}
\end{table}

\begin{table}[t]
    \centering
    \caption{Long-context retrieval results (MLDR), reported as nDCG@10 at varying maximum token lengths. Best results in \textbf{bold}, second best \underline{underlined}. Models limited in context tokens are marked with --.}
    \label{tab:mldr_results}
    \begin{tabular}{lcccc}
        \toprule
        \textbf{Model} & \textbf{512} & \textbf{2048} & \textbf{4096} & \textbf{8192} \\
        \midrule
        \multicolumn{5}{l}{\textit{Baselines}} \\
        BERT-base         & 0.4029 & --     & --     & --     \\
        BERTimbau-base    & 0.5119 & --     & --     & --     \\
        ModernBERT-base   & 0.4054 & 0.4206 & 0.3015 & 0.2867 \\
        NeoBERT-base      & 0.4746 & 0.5149 & 0.4676 & --     \\
        Qwen-base         & 0.3560 & 0.4023 & 0.4241 & 0.5351 \\
        \midrule
        \multicolumn{5}{l}{\textit{Our models}} \\
        moBERTo (orig.\ tok.)      & \textbf{0.5834} & \underline{0.5909} & \textbf{0.6286} & \textbf{0.6166} \\
        moBERTo-8k (orig.\ tok.)   & 0.5674 & \textbf{0.6025} & 0.5876 & \underline{0.6140} \\
        moBERTo-SWM (PT tok.)      & 0.5466 & 0.4791 & 0.5714 & 0.5857 \\
        moBERTo-SWM-8k (PT tok.)   & \underline{0.5827} & 0.5606 & \underline{0.5905} & 0.5777 \\
        \midrule
        \multicolumn{5}{l}{\textit{Other architectures}} \\
        NeoBERT-PT           & 0.4396 & 0.4030 & 0.4328 & --     \\
        Qwen3-0.6B-PT              & 0.3873 & 0.3413 & 0.3420 & 0.4916 \\
        \bottomrule
    \end{tabular}
\end{table}

\begin{table}[t]
    \centering
    \caption{Classification results, reported as F1. Best results in \textbf{bold}, second best \underline{underlined}.}
    \label{tab:cls_results}
    \begin{tabular}{lccc}
        \toprule
        \textbf{Model} & \textbf{Docs} & \textbf{Educ.} & \textbf{Avg.} \\
        \midrule
        \multicolumn{4}{l}{\textit{Baselines}} \\
        BERT-base         & 0.8700 & 0.5690 & 0.7195 \\
        BERTimbau-base    & 0.8978 & 0.6382 & 0.7680 \\
        ModernBERT-base   & 0.8416 & 0.5730 & 0.7073 \\
        NeoBERT-base      & 0.8970 & 0.6266 & 0.7618 \\
        Qwen-base         & \textbf{0.9120} & 0.6289 & 0.7705 \\
        \midrule
        \multicolumn{4}{l}{\textit{Our models}} \\
        moBERTo (orig.\ tok.)      & 0.8942 & 0.6070 & 0.7506 \\
        moBERTo-8k (orig.\ tok.)   & 0.8962 & 0.6035 & 0.7499 \\
        moBERTo-SWM (PT tok.)      & 0.9024 & 0.6281 & 0.7653 \\
        moBERTo-SWM-8k (PT tok.)   & 0.9039 & \underline{0.6394} & \underline{0.7717} \\
        \midrule
        \multicolumn{4}{l}{\textit{Other architectures}} \\
        NeoBERT-PT           & 0.9030 & \textbf{0.6428} & \textbf{0.7729} \\
        Qwen3-0.6B-PT              & \underline{0.9070} & 0.6311 & 0.7691 \\
        \bottomrule
    \end{tabular}
\end{table}

\begin{table}[t]
    \centering
    \caption{NLU, NER, and English (GLUE) results. Best results in \textbf{bold}, second best \underline{underlined}. GLUE measures the trade-off between Portuguese adaptation and English retention.}
    \label{tab:nlu_ner_results}
    \begin{tabular}{lccc}
        \toprule
        \textbf{Model} & \textbf{GLUE} & \textbf{PLUE-PT} & \textbf{LeNER-br} \\
        \midrule
        \multicolumn{4}{l}{\textit{Baselines}} \\
        BERT-base         & \underline{0.7815} & 0.6423 & 0.8500 \\
        BERTimbau-base    & 0.6772 & 0.6800 & \textbf{0.9040} \\
        ModernBERT-base   & \textbf{0.8301} & 0.6420 & 0.8240 \\
        NeoBERT-base      & 0.7430 & 0.6654 & 0.8590 \\
        Qwen-base         & 0.7260 & 0.6343 & 0.7020 \\
        \midrule
        \multicolumn{4}{l}{\textit{Our models}} \\
        moBERTo (orig.\ tok.)      & 0.7705 & 0.6849 & 0.8371 \\
        moBERTo-8k (orig.\ tok.)   & 0.7724 & 0.6910 & 0.8587 \\
        moBERTo-SWM (PT tok.)      & 0.7128 & \underline{0.6959} & 0.8710 \\
        moBERTo-SWM-8k (PT tok.)   & 0.7354 & \textbf{0.6980} & 0.8726 \\
        \midrule
        \multicolumn{4}{l}{\textit{Other architectures}} \\
        NeoBERT-PT           & 0.6620 & 0.6842 & \underline{0.8840} \\
        Qwen3-0.6B-PT              & 0.7050 & 0.6632 & 0.7100 \\
        \bottomrule
    \end{tabular}
\end{table}

\subsection{Information Retrieval}

Table~\ref{tab:ir_results} reports reranking performance as nDCG@10. 
moBERTo-SWM-8k a\-chieves the highest average (0.5255), followed by 
moBERTo-SWM (0.5120) and moBERTo (orig.\ tok.) (0.5001), all above 
the strongest baseline, BERTimbau-base (0.4671). moBERTo-SWM-8k 
ranks first on QUATI (0.5609) and Robust04-PT (0.5010), while 
moBERTo-SWM leads on mMARCO-PT (0.5169), suggesting that 
subword-matching embedding transfer and long-context post-training 
contribute complementary improvements. Continued pretraining on 
Portuguese yields average gains for all moBERTo variants over 
ModernBERT-base, but the benefit is not uniform across architectures: 
NeoBERT-PT shows only a marginal improvement over NeoBERT-base, and 
Qwen3-0.6B-PT, while improving on average, degrades on Robust04-PT 
relative to Qwen3-0.6B-base.

\subsection{Long-Context Retrieval (MLDR)}

Long-context retrieval results (Table~\ref{tab:mldr_results}) reveal 
two main findings. First, mo\-BERTo (orig.\ tok.) achieves the strongest 
results at 512 (0.5834), 4{,}096 (0.6286), and 8{,}192 tokens (0.6166), 
and the second best at 2{,}048 tokens (0.5909), despite being trained 
exclusively at 1{,}024 tokens. This supports our hypothesis that the 
long-context capabilities of ModernBERT transfer effectively through 
continued pretraining without a dedicated long-context training stage. 
Second, the dedicated long-context post-training phase yields mixed 
effects: moBERTo-8k improves over moBERTo (orig.\ tok.) at 2{,}048 
tokens (0.6025 vs.\ 0.5909) but slightly underperforms it at 4{,}096 
(0.5876 vs.\ 0.6286) and 8{,}192 tokens (0.6140 vs.\ 0.6166), 
suggesting that the additional 10B-token phase does not consistently 
improve long-context performance when starting from an already strong 
base. The SWM variants follow a similar pattern: moBERTo-SWM-8k 
improves over moBERTo-SWM at 512 (0.5827 vs.\ 0.5466) and 4{,}096 
tokens (0.5905 vs.\ 0.5714), but is slightly lower at 8{,}192 tokens 
(0.5777 vs.\ 0.5857). In contrast, ModernBERT-base degrades 
substantially beyond 2{,}048 tokens on Portuguese (from 0.4054 at 512 
to 0.2867 at 8{,}192), confirming that language-specific adaptation is 
necessary for long-context performance even when architectural support 
is already present. The impact of tokenizer adaptation on long-context 
performance is analyzed in section~\ref{sec:ablations}.

\subsection{Classification}

Table~\ref{tab:cls_results} reports classification performance as F1. 
On the document type classification (Docs) and educational content 
classification (Educ.) benchmarks, we observe relatively small 
variation across models. NeoBERT-PT achieves the highest average 
(0.7729), followed by moBERTo-SWM-8k (0.7717) and Qwen3-0.6B-PT 
(0.7691). Notably, even the English-only BERT-base achieves 0.87 on 
Docs, indicating that the task may not require deep language-specific 
understanding. The SWM variants consistently outperform the 
original-tokenizer variants on classification (e.g., moBERTo-SWM-8k 
0.7717 vs.\ moBERTo-8k 0.7499), suggesting that Portuguese-specific 
tokenization benefits these tasks. We interpret these benchmarks as 
less discriminative overall for evaluating the effect of language 
adaptation.

\subsection{NLU and NER}

Table~\ref{tab:nlu_ner_results} reports results on PLUE-PT, LeNER-Br, 
and GLUE. On PLUE-PT, the SWM variants achieve the strongest results 
among all models, with moBERTo-SWM-8k leading at 0.6980, above 
moBERTo (orig.\ tok.) and BERTimbau-base. This suggests that 
Portuguese-specific tokenization combined with embedding transfer is 
beneficial for natural language understanding. On LeNER-Br, 
BER\-Timbau achieves the best result (0.9040), followed by NeoBERT-PT (0.8840). Among 
moBERTo variants, the SWM models achieve the strongest NER results, 
with moBERTo-SWM-8k reaching 0.8726, substantially above moBERTo 
(orig.\ tok.), indicating that Portuguese tokenization and embedding 
transfer help on token-level tasks. The long-context variants also 
improve over their respective base counterparts, suggesting that 
longer-context training benefits NER as well. As expected, continued 
pretraining on Portuguese degrades English performance: ModernBERT-base 
leads on GLUE (0.8301), while all moBERTo variants drop, with the SWM 
variants showing the largest decrease due to the new tokenizer.

\section{Ablation Studies}
\label{sec:ablations}

\begin{table}[t]
    \centering
    \caption{Ablation results across all design decisions. Reranking 
    Avg.\ is the mean of QUATI, mMARCO-PT, and Robust04-PT. 
    Classification Avg.\ is the mean of Docs and Educ. All Portuguese 
    adaptations are trained on the same 60B-token corpus.}
    \label{tab:ablations}
    \setlength{\tabcolsep}{4.9pt}
    \begin{tabular}{lcccccc}
        \toprule
        \textbf{Variant} & \textbf{Rerank} & \textbf{MLDR} & \textbf{Class.} & \textbf{PLUE} & \textbf{LeNER} & \textbf{GLUE} \\
         & \textbf{Avg.} & \textbf{8{,}192} & \textbf{Avg.} & \textbf{-PT} & \textbf{-Br} & \\
        \midrule
        \multicolumn{7}{l}{\textit{Initialization (ModernBERT-base, orig.\ tok.)}} \\
        moBERTo (orig.\ tok.)   & 0.5001 & \textbf{0.6166} & 0.7506 & 0.6849 & 0.8371 & 0.7705 \\
        moBERTo (scratch)       & 0.4524 & 0.1405          & 0.7726 & 0.6700 & 0.8500 & 0.7130 \\
        \midrule
        \multicolumn{7}{l}{\textit{Tokenizer adaptation}} \\
        moBERTo-Tok             & 0.4961 & 0.5036 & 0.7557 & 0.7119 & 0.8435 & 0.7066 \\
        moBERTo-SWM             & 0.5120 & 0.5857 & 0.7653 & 0.6959 & 0.8710 & 0.7128 \\
        \midrule
        \multicolumn{7}{l}{\textit{Long-context post-training (+10B tokens at 8{,}192)}} \\
        moBERTo-8k              & 0.4945 & 0.6140 & 0.7499 & 0.6910          & 0.8587          & \textbf{0.7724} \\
        moBERTo-SWM-8k          & \textbf{0.5255} & 0.5777 & \textbf{0.7717} & \textbf{0.6980} & \textbf{0.8726} & 0.7354 \\
        \midrule
        \multicolumn{7}{l}{\textit{Base architecture (continued pretraining on the same 60B-token corpus)}} \\
        NeoBERT-PT              & 0.3964 & --     & \textbf{0.7729} & 0.6842 & \textbf{0.8840} & 0.6620 \\
        Qwen3-0.6B-PT           & 0.4195 & 0.4916 & 0.7691 & 0.6632 & 0.7100 & 0.7050 \\
        \bottomrule
    \end{tabular}
\end{table}

We isolate the effect of four design decisions: initialization 
(continued pretraining vs.\ from scratch), tokenizer adaptation 
(original ModernBERT tokenizer vs.\ a Portuguese tokenizer with 
or without SWM embedding transfer), long-context post-training, 
and the choice of base architecture. All variants are trained on 
the same 60B-token Portuguese corpus, with optional 10B-token 
post-training at 8{,}192 tokens for the -8k variants. 
Table~\ref{tab:ablations} summarizes results.

\subsubsection*{Initialization.}
Continued pretraining outperforms training from scratch on retrieval 
(+4.8 nDCG@10 on average) and especially on long-context retrieval 
(+48 points on MLDR@8{,}192). Notably, both variants are trained 
under the same 1{,}024-token Portuguese budget; the gap on long-context 
retrieval reflects the fact that moBERTo (orig.\ tok.) inherits 
long-context representations from the original 2T-token ModernBERT 
pretraining, while moBERTo (scratch) has no such prior exposure to 
long sequences. On classification, the from-scratch model performs 
comparably, consistent with these tasks being near saturation 
(Sect.~\ref{sec:results}).

\subsubsection*{Tokenizer adaptation.}

A Portuguese tokenizer benefits token-level tasks, with both 
moBERTo-Tok and moBERTo-SWM improving over the base on PLUE-PT and 
LeNER-Br. However, replacing the tokenizer disrupts long-context 
retrieval: moBERTo-Tok drops substantially on MLDR@8{,}192 (0.5036 
vs.\ 0.6166 for the base). SWM embedding transfer mitigates this 
loss (0.5857), since initializing each new token's embedding from 
the original subword embeddings keeps the model close to its 
pretrained representation space and preserves the position-content 
mapping required for long sequences.

\subsubsection*{Long-context post-training.}

An additional 10B tokens at 8{,}192-token context yields the strongest 
reranking model overall (moBERTo-SWM-8k, 0.5255 nDCG@10), with the 
largest gains on Robust04-PT (+2.3) and consistent improvements on 
NER. The benefit is not uniform on MLDR itself, where the post-trained 
variants are slightly below their 1{,}024-context counterparts at 
8{,}192 tokens, suggesting that further long-context gains are 
bottlenecked by factors beyond sequence length exposure.

\subsubsection*{Base architecture.}

We additionally adapt NeoBERT-base and Qwen3-0.6B-base under the 
same recipe (NeoBERT-PT, Qwen3-0.6B-PT) to test whether the gains 
observed for ModernBERT-PT are tied to the architecture or simply 
follow from continued pretraining on a large Portuguese corpus. 
On reranking, ModernBERT-PT substantially outperforms both 
alternatives (0.5001 vs.\ 0.3964 for NeoBERT-PT and 0.4195 for 
Qwen3-0.6B-PT), despite Qwen3-0.6B having roughly four times more 
parameters. This indicates that scaling parameter count alone, 
without architectural features tailored to bidirectional encoding 
and long contexts, does not close the gap on retrieval for this range of parameters.

The two baselines fail in different ways. NeoBERT-PT remains 
competitive on classification (0.7729 Class.\ Avg.) and achieves 
the strongest LeNER-Br score (0.8840), but trails on reranking 
and lacks native long-context support. Qwen3-0.6B-PT handles 
long contexts (0.4916 on MLDR@8{,}192) but underperforms on 
shorter-context tasks, particularly NER (--16 points on LeNER-Br).

\section{Conclusion}

We presented moBERTo, a Portuguese adaptation of ModernBERT obtained through continued pretraining on a curated 60-billion-token corpus. Our model brings modern encoder advances, including Rotary Positional Embeddings, alternating attention, Flash Attention, and unpadding. Our best variant, moBERTo-SWM-8k, which combines a Portuguese tokenizer with subword-matching embedding transfer and long-context post-training, achieves the highest average reranking nDCG@10 (0.5255) across three Portuguese retrieval benchmarks, outperforming BERTimbau (0.4671) and all other baselines. On long-context retrieval, moBERTo with the original tokenizer achieves the strongest results at 8{,}192 tokens (0.6166) despite being trained exclusively at 1{,}024 tokens.

Our ablation studies reveal that (i) continued pretraining is strongly preferable to training from scratch, especially for long-context capabilities; (ii) tokenizer adaptation benefits token-level tasks but degrades long-context retrieval due to positional misalignment; (iii) SWM embedding transfer mitigates this degradation while improving reranking and NER; and (iv) a dedicated long-context post-training phase provides further gains on reranking and NER. We publicly release the model weights and training data to support future research and applications in Portuguese NLP.

\subsubsection*{Acknowledgments.}
We thank Maritaca AI for providing the computational infrastructure used 
to train and evaluate the models presented in this work.
%
%
\bibliographystyle{splncs04}
\bibliography{references}

\end{document}